\documentclass[runningheads]{llncs}

\usepackage[mobile]{eccv}

\PassOptionsToPackage{pagebackref}{hyperref}

\newcommand{\tablestyle}[2]{\setlength{\tabcolsep}{#1}\renewcommand{\arraystretch}{#2}\centering\footnotesize}

\newcommand{\ours}{SelfEvo\xspace}

\usepackage{eccvabbrv}

\usepackage{graphicx}
\usepackage{booktabs}

\usepackage[accsupp]{axessibility}  %

\usepackage{multirow}
\usepackage{makecell} 
\usepackage{arydshln}
\usepackage{booktabs}

\usepackage{hyperref}

\usepackage{orcidlink}

\begin{document}

\title{Self-Improving 4D Perception via Self-Distillation}

\titlerunning{SelfEvo: Self-Improving 4D Perception via Self-Distillation}

\author{
    Nan Huang$^{1,2*}$ \quad
    Pengcheng Yu$^{2,4*}$ \quad
    Weijia Zeng$^{6}$ \quad
    James M. Rehg$^{1}$ \\[0.2em]
    Angjoo Kanazawa$^{5}$ \quad
    Haiwen Feng$^{2,5\dagger}$ \quad
    Qianqian Wang$^{2,3\dagger}$
}
\institute{
    $^1$UIUC \enspace
    $^2$Impossible Research \enspace
    $^3$Harvard \\[0.1em]
    $^4$MPI for Intelligent Systems \enspace
    $^5$UC Berkeley \enspace
    $^6$UBC
}
\authorrunning{N. Huang et al.}

\setlength{\footnotesep}{1.2em}        %
\setlength{\skip\footins}{2.0em}        %
\maketitle

\renewcommand{\thefootnote}{}
\footnotesize{\footnotetext{$^*$Equal contribution. $^\dagger$Equal advising.}}
\renewcommand{\thefootnote}{\arabic{footnote}}
\setcounter{footnote}{0}

{
\centering
\includegraphics[width=1.0\textwidth]{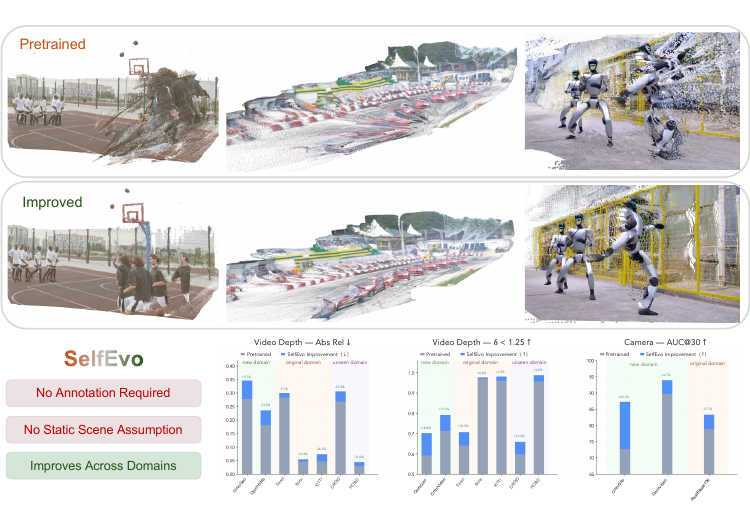}
\vspace{-0.5em}
{\small
\captionsetup{hypcap=false}\captionof{figure}{
We present \ours{}, a self-improving framework for learning-based multi-view reconstruction via self-distillation, requiring no ground-truth annotations.
}
}

\label{fig:teaser}
}

\begin{abstract}
Large-scale multi-view reconstruction models have made remarkable progress, but most existing approaches still rely on fully supervised training with ground-truth 3D/4D annotations. Such annotations are expensive and particularly scarce for dynamic scenes, limiting scalability. We propose \textbf{\ours}, a self-improving framework that continually improves pretrained multi-view reconstruction models using unlabeled videos. \ours introduces a self-distillation scheme using spatiotemporal context asymmetry, enabling self-improvement for learning-based 4D perception without external annotations. We systematically study design choices that make self-improvement effective, including loss signals, forms of asymmetry, and other training strategies. Across eight benchmarks spanning diverse datasets and domains, \ours consistently improves pretrained baselines and generalizes across base models (e.g. VGGT and $\pi^3$), with significant gains on dynamic scenes. Overall, \ours achieves up to 36.5\% relative improvement in video depth estimation and 20.1\% in camera estimation, without using any labeled data. Project Page: \url{https://self-evo.github.io/}.

\end{abstract}
    
\section{Introduction}

Recent learning-based multi-view reconstruction models such as DUSt3R~\cite{wang2024dust3r}, VGGT~\cite{wang2025vggt}, $\pi^3$~\cite{wang2025pi}, and DA3~\cite{depthanything3} demonstrate that large-scale supervised training can imbue networks with strong geometric priors, enabling robust feedforward 3D prediction even in under-constrained settings. Despite their effectiveness, these methods rely on dense ground-truth geometric annotations for fully supervised training, a form of supervision that is costly to acquire even for static scenes, difficult to scale to in-the-wild scenarios, and increasingly prohibitive for dynamic (4D) settings where annotated data is scarce. Consequently, current systems are constrained not only by the lack of large-scale 3D/4D annotations, but also by a rigid train-freeze-deploy paradigm in which models are trained once on curated datasets and remain fixed at deployment, with no mechanism to adapt when the target data distribution shifts.

In this work, we ask: can multi-view reconstruction models continually improve on unlabeled data, without requiring any geometric annotations?
To this end, we propose \textbf{\ours}~(\textbf{Self}-\textbf{Evo}lving 4D Perception), a framework that enables self-improving 4D perception via \textit{self-distillation}. Our key insight stems from a simple observation: multi-view reconstruction models produce more reliable geometric predictions when provided with richer spatiotemporal context (\ie, denser input views) than when operating on sparser inputs. This motivates the core idea: even without supervision, the model can continuously learn from itself through self-distillation by leveraging asymmetric spatiotemporal context.

Specifically, our self-improvement framework works as follows. Starting from a pretrained model, we instantiate two instances, a teacher and a student. During training, the teacher receives a richer set of input views, while the student is restricted to a subset. Predictions from the richer-context teacher are used as pseudo-targets to supervise the student. The teacher is maintained as an exponential moving average of the student, forming an online self-improving loop that continuously bootstraps better 4D predictions from unlabeled videos. In this way, \textit{spatiotemporal context asymmetry} serves as an effective self-supervision signal, enabling the model to continually improve its 4D perception without any annotations.

Beyond proposing a new framework, we also systematically study design choices for effective self-improvement of multi-view reconstruction models, including spatiotemporal context asymmetry construction, loss signals, and training strategies. Some of our findings include (i) frame dropping is the most effective way to introduce the information asymmetry, (ii) an online self-improving loop outperforms fixed-teacher training, and (iii) selectively freezing the camera decoder further improves performance. The complete analysis can be found in Sec.~\ref{sec:analysis}.

We evaluate our framework across multiple base models (e.g., VGGT~\cite{wang2025vggt} and $\pi^3$~\cite{wang2025pi}) and target domains (e.g., OmniWorld-Game~\cite{zhou2025omniworld} and BEDLAM2.0~\cite{tesch2025bedlam2}). The resulting models achieve significant gains on the target domains and also improve performance on related unseen domain datasets such as DROID~\cite{khazatsky2024droid} and HOI4D~\cite{Liu_2022_CVPR}. Importantly, performance on the original domains is preserved and often further improved, including on the native evaluation benchmarks of VGGT and $\pi^3$. Compared with a supervised fine-tuning (SFT) baseline that relies on ground-truth annotations, our method requires no annotations yet exhibits stronger cross-domain generalization while better preserving performance on the original domain. 

In summary, our key contributions are: (1) a novel self-improving framework for learning-based multi-view reconstruction based on context asymmetry; (2) a comprehensive and systematic study of design choices for effective self-improvement within this framework; and (3) strong empirical results across models and domains, demonstrating improved in-domain performance, stronger cross-domain generalization, and better retention on the original domains.

\section{Related Work}
\label{sec:related}

\noindent \textbf{Feedforward Multi-View Reconstruction.} 3D reconstruction~\cite{hartley2003multiple} has  been a central problem in computer vision. Classical methods~\cite{schoenberger2016sfm,agarwal2011building,snavely2006photo,wu2013towards,schoenberger2016mvs,furukawa2015multi,schonberger2016pixelwise} recover geometry and cameras through hand-crafted optimization pipelines. With deep learning, many components of these pipelines, such as feature detection and matching~\cite{lowe1999object,bay2006surf}, have been replaced by learned modules~~\cite{detone2018superpoint,yi2016lift,lindenberger2023lightglue,sarlin2020superglue,yao2018mvsnet,dusmanu2019d2}. 
More recently, DUSt3R~\cite{wang2024dust3r} demonstrated the potential of large-scale transformers~\cite{vaswani2017attention,dosovitskiy2020image} to directly infer dense geometry and camera parameterrs from image pairs, marking the beginning of a new paradigm~\cite{leroy2024grounding,zhang2024monst3r,chen2025long3r,murai2025mast3r,lu2025align3r,duisterhof2025mast3r} for unified, feedforward 3D reconstruction. Successors~\cite{wang2025pi,yang2025fast3r,zhang2025flare,tang2024mv, depthanything3} such as VGGT~\cite{wang2025vggt}, CUT3R~\cite{wang2025continuous} and $\pi^3$~\cite{wang2025pi}, extend this framework to fully feedforward \emph{multi-view} reconstruction, highlighting the capability of large-scale models to handle sparse and unconstrained settings. However, these methods remain strongly supervised, relying on large-scale ground-truth geometric annotations, which limits their scalability, particularly for dynamic scenes where annotations are harder to obtain.

\vspace{0.5em}
\noindent \textbf{Self-Supervised 3D Learning.} 
Prior work has explored self supervised learning of 3D structure from unlabeled videos. A dominant approach leverages photometric consistency~\cite{zhou2017unsupervised,godard2019digging,bian2019unsupervised,mahjourian2018unsupervised,yang2020d3vo,lai2021video,fu2022mononerf} to jointly learn geometry and camera motion. However, photometric consistency becomes unreliable under large viewpoint changes, dynamic scenes, or strong view-dependent effects. As a result, many methods either perform well only on continuous, mostly static video streams~\cite{zhou2017unsupervised,bian2019unsupervised,lai2021video,zhao2025rayzer,jiang2025rayzer,sajjadi2022rust} or operate under category-specific constraints~\cite{chanmonteiro2020pi-GAN,cmrKanazawa18,lin2020sdfsrn,mustikovelaCVPR20,HoloGAN2019,yan2016perspective,schwarz2020graf}. Another line of work learns representations rather than explicit geometry. For example, CroCo-v2~\cite{croco,croco_v2} pretrains representations using a pairwise masked autoencoding framework for downstream geometric tasks~\cite{wang2024dust3r}. In contrast to these methods, we explore self-improvement of geometric foundation models after pretraining. Starting from pretrained models trained with 3D supervision, we further improve them using unlabeled videos. This leverages existing supervision while enabling continued improvement with unlabeled data, pushing performance beyond the original models on new domains and in-the-wild videos. In addition, our training signal comes from a self-distillation formulation rather than multi-view consistency, making it more general and scalable, and enabling training on in-the-wild dynamic scenes.

\vspace{0.5em}
\noindent \textbf{Self-Training and Knowledge Distillation.} Our method is broadly related to knowledge distillation~\cite{hinton2015distilling,buciluǎ2006model,xie2020self,tarvainen2017mean,chen2020big,zhang2019your,touvron2021training}. Unlike traditional formulations, where teacher and student receive identical inputs and knowledge is distilled from a stronger teacher into a smaller student, 
our framework explores self-distillation~\cite{chefer2026self,shenfeld2026self,song2026expanding,zhang2026embarrassingly} in which the teacher and student share the same architecture and initialization, while  spatiotemporal context asymmetry enables self-improvement as the two networks co-evolve.
Similar ideas have been explored in representation learning (e.g., BYOL~\cite{doersch2024bootstap} and DINO~\cite{caron2021emerging,oquab2023dinov2}) and in methods that bootstrap existing vision models. For example, in monocular depth estimation, Depth Anything~\cite{yang2024depth} bootstraps a model pretrained on synthetic data using unlabeled real-world images, where the student’s input is perturbed with color jittering and CutMix~\cite{yun2019cutmix}. In 2D point tracking, BootsTAP~\cite{doersch2024bootstap} adopts a similar teacher-student framework, corrupting the student’s input through color jittering and random cropping.

However, such approaches remain relatively unexplored for multi-view reconstruction models.
These models are naturally flexible, as they can process both continuous videos and unstructured photo collections, making them well suited for constructing context asymmetry during training and offering a broad design space.
Concurrent with our work, Selfi~\cite{deng2025selfi} proposes a self-improving pipeline built on {geometric feature alignment}. 
It freezes a 3D foundation model and trains a lightweight feature adapter using reprojection-based feature consistency. However, this design is limited to static scenes due to the underlying multi-view consistency assumption. In contrast, our method enables online continual self-improvement of the reconstruction model through self-distillation without requiring scenes to be static, making it more general and scalable.

\section{Method}
\label{sec:method}

In Sec.~\ref{sec:pre}, we discuss key properties of learning-based multi-view reconstruction that motivate our framework. In Sec.~\ref{sec:sica}, we introduce our framework \ours{} and define the design space. We summarize the default instantiation of the framework there, with ablations and analysis deferred to Sec.~\ref{sec:analysis}.

\subsection{Preliminaries}
\label{sec:pre}
Learning-based multi-view reconstruction models such as VGGT~\cite{wang2025vggt} and $\pi^3$~\cite{wang2025pi} take a set of images $I_i \in \mathcal{I}$ as input and predict the geometric outputs for each frame $O_i \in \mathcal{O}$ using a transformer-based architecture. Each $O_i$ includes the camera parameters and dense geometry under various parameterizations. During training, these models are presented with diverse combinations of input views and scene types. 

Despite this flexibility, the quality of the feedforward predictions still depends strongly on the amount of contextual information available where providing more views tends to lead to better reconstruction. 
We empirically verify this trend
in the supplementary material~(Sec.~\ref{sec:supp_pre}). 
This observation motivates our self-distillation framework, which converts spatiotemporal context asymmetry into a supervision signal that drives self-improvement.

\begin{figure*}[t]
    \centering
    \includegraphics[width=\textwidth]{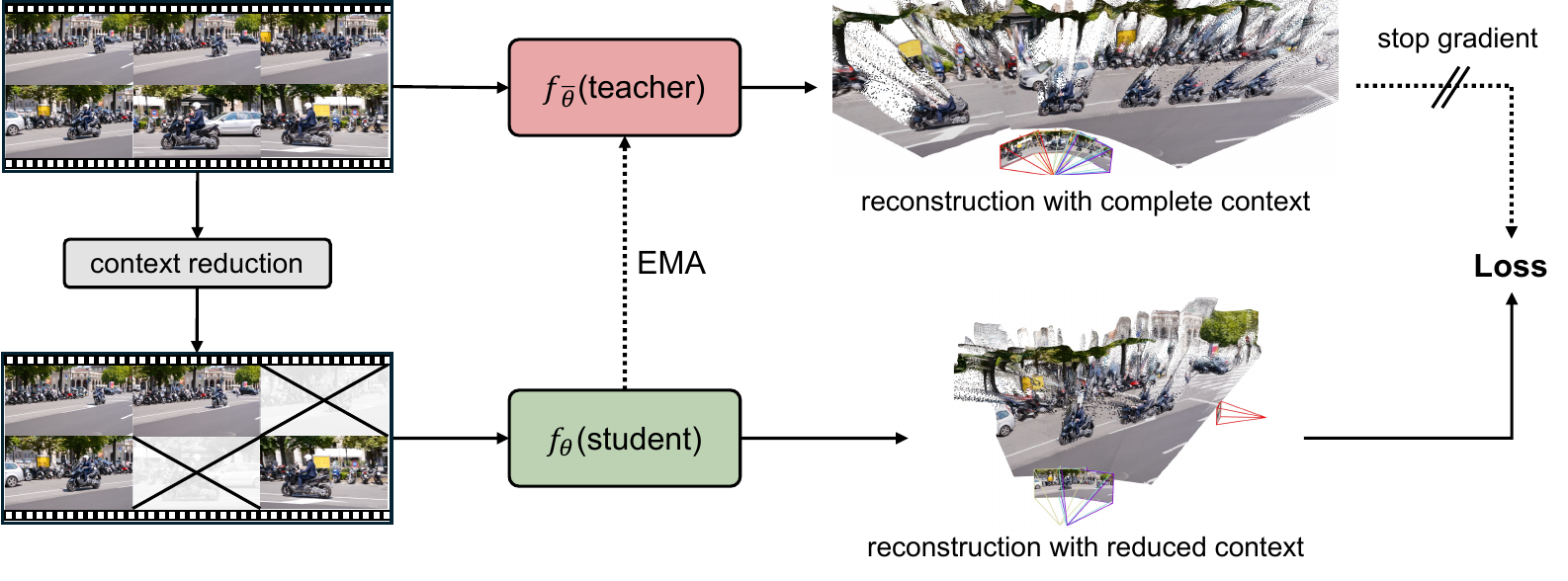}
    \caption{
    We propose an annotation-free self-improving framework that continually post-trains pretrained multi-view reconstruction models using unlabeled videos. Our method forms an online self-distillation loop where a richer-context teacher provides stop-gradient pseudo targets to a student operating on reduced context, and is updated as an EMA of the student after each step.
    }
    \label{fig:placeholder}
\end{figure*}

\subsection{The \ours\ Framework}
\label{sec:sica}
We now introduce
\ours, an annotation-free self-improving framework for multi-view reconstruction models. 
It exploits spatiotemporal context asymmetry: predictions from richer spatiotemporal context are typically more reliable than those from restricted context,
as broader-view sequences often contain stronger multi-view constraints that these models can leverage.
We instantiate this idea with a self-distillation loop: the teacher observes higher-context inputs to produce pseudo targets, while the student is provided with reduced context  and learns to match teacher's outputs.

\vspace{0.5mm}
\noindent\textbf{Setup.}
Given an unlabeled clip $x=\{I_t\}_{t=1}^{S}$, we construct the input for the teacher and the student as:
\begin{equation}
x_T = a_T(x), \qquad x_S = a_S(x_T;\beta),
\label{eq:views}
\end{equation}
where $a_T(\cdot)$ forms a context-rich input and $a_S(\cdot)$ reduces context to induce asymmetry, and $\beta$ specifies how the student context is reduced.
Let $f_{\bar\theta}$ and $f_{\theta}$ denote teacher and student models (same architecture, same initialization). We compute
\begin{equation}
O^T = f_{\bar\theta}(x_T), \qquad O^S = f_{\theta}(x_S).
\label{eq:pred}
\end{equation}

\noindent\textbf{Objective.}
We optimize the student parameters $\theta$ with gradient descent:
\begin{equation}
\min_{\theta}\;\mathcal{L}(\theta;\beta,\pi,\tau,\gamma)
=
\underbrace{\mathcal{L}_{\text{base}}\!\left(O^S,\ \mathrm{sg}[O^T]\right)}_{\text{output-level self-distillation}}
+\gamma\underbrace{\mathcal{L}_{\text{feat}}\!\left(F^S,\ \mathrm{sg}[F^T]\right)}_{\text{optional feature matching}}.
\label{eq:design_objective}
\end{equation}
Here $\mathrm{sg}[\cdot]$ denotes stop-gradient, and $F^S$ and $F^T$ represent intermediate feature representations from the student and teacher models, respectively. $\pi$ specifies which parameters are trainable (others are frozen), $\tau$ determines whether the teacher is fixed or updated online, and $\gamma\ge0$ weights the feature matching loss.
In the online setting, the teacher is commonly updated via EMA:
\begin{equation}
\bar{\theta} \leftarrow \lambda \bar{\theta} + (1-\lambda)\theta.
\label{eq:ema}
\end{equation}

\vspace{0.5em}
\noindent\textbf{Design axes.}
Eq.~\eqref{eq:design_objective} defines a  design space for context-asymmetric self-improvement. 
We investigate five design axes: how to construct teacher/student inputs, how to select student frames, how to update the teacher, which parameters to update, and whether to incorporate feature-level supervision. We summarize the main findings below; the corresponding ablations and analysis are reported in Sec.~\ref{sec:analysis}.

\vspace{0.5em}
\noindent\textbf{(a) Inducing context asymmetry $(a_T,a_S)$.}
We compare three ways to induce context asymmetry: photometric perturbations, frame cropping~(spatially cropping input frames), and frame dropping~(keeping only a subset of input frames). Across all settings, frame dropping proves most effective~(\S\ref{sec:ablation_asymmetry}, Tab.~\ref{tab:ablation_asymmetry_omnigeo_omnivideo}).

\vspace{0.5em}
\noindent\textbf{(b) Student frame selection (scheme $\beta$ inside $a_S$).} Under frame dropping, the student observes a subsequence selected from the teacher clip.
We consider random sampling and attention guided sampling, where frame importance is derived from teacher attention scores.
Overall, random sampling is the most robust and performs best across settings (\S\ref{sec:frame_selection}, Tab.~\ref{tab:ablation_frame_sample}).

\vspace{0.5em}
\noindent\textbf{(c) Teacher update rule $\tau$.}
We compare a fixed teacher ($\lambda=1$) with an online teacher that co-evolves with the student~($0 < \lambda < 1$).
An online updated teacher consistently outperforms offline pseudo-label fine-tuning (\S\ref{sec:online_vs_offline}, Tab.~\ref{tab:ablation_online}).

\vspace{0.5em}
\noindent\textbf{(d) Parameter update rule $\pi$.}
To balance adaptation and stability under imperfect pseudo supervision, we explore freezing different components (camera decoder, depth decoder, backbone, and combinations). Freezing the camera decoder while updating the rest yields the best results (\S\ref{sec:what_to_update}, Tab.~\ref{tab:ablation_training_recipe}).

\vspace{0.5em}
\noindent\textbf{(e) Supervision form.}
Beyond output-level distillation, we explore adding intermediate feature matching loss $\mathcal{L}_{\text{feat}}$, which does not yield significant gains~(\S\ref{sec:loss_design}, Tab.~\ref{tab:ablation_loss}).

\vspace{0.5em}
\noindent\textbf{Default instantiation.}
Unless stated otherwise, we use random frame dropping, an online EMA teacher updated every step, freeze the camera decoder, and only apply output-level self-distillation ($\gamma{=}0$).

\section{Experiments}
\label{sec:exp}

We first describe the experimental setup in Sec.~\ref{sec:setup}. We then report our main self-improvement results on the primary setting (VGGT self-improved on OmniWorld-Game) in Sec.~\ref{sec:cl_eval}.
In Sec.~\ref{sec:cl_generality}, we demonstrate the generality of our framework across base models and training sources.
Finally, Sec.~\ref{sec:ood_generalization} evaluates unseen-domain generalization beyond the adaptation domain.

\subsection{Experimental Setup}
\label{sec:setup}
\noindent \textbf{Base Models.}
We evaluate our self-improvement framework on two models, VGGT~\cite{wang2025vggt} and $\pi^3$~\cite{wang2025pi}, to assess its applicability across different architectures and training objectives. During self-improvement, we retain the original training losses of each model while replacing ground-truth annotations with teacher-generated pseudo labels. For VGGT we optimize the camera and depth losses, and for $\pi^3$ we optimize the camera and point-map losses.

\vspace{0.5em}
\noindent \textbf{Training Data.}
For VGGT, we perform self-improvement on OmniWorld-Game~\cite{zhou2025omniworld}, a large-scale synthetic video dataset with diverse game environments. Note that our framework requires no labels, we do not use any annotations from OmniWorld-Game during training. 
We use OmniWorld-Game mainly for evaluation purposes, as it provides ground-truth annotations for quantitative assessment.
To verify the generality, we additionally evaluate our framework on  {both} VGGT and $\pi^3$ using BEDLAM2.0~\cite{tesch2025bedlam2} and DROID~\cite{khazatsky2024droid} datasets. 
BEDLAM2.0 is a large-scale synthetic human-centric video dataset and
DROID is a real-world robot manipulation dataset with markedly different scene layouts and motion patterns. 
In all settings, we use only RGB frames during self-improvement and do not use any dataset-provided geometry annotations.

\vspace{0.5em}
\noindent \textbf{Benchmark.}
Driven by the goal of continual self-improvement, our evaluation encompasses two key dimensions: \emph{new-domain adaptation} and \emph{original-domain retention}. 
For new domain adaptation, we use in-distribution benchmarks that match the unlabeled training distribution. When an official benchmark is available (OmniWorld-Game~\cite{zhou2025omniworld}), we follow the released protocol and evaluate on Game-Benchmark. For other training sources (BEDLAM2.0 and DROID), we construct in-distribution benchmarks by randomly sampling held-out subsets from the same datasets.
Game-Benchmark contains two subsets, OmniGeo and OmniVideo. OmniGeo emphasizes challenging geometric understanding with larger camera/object motion, while OmniVideo features complex camera trajectories with generally milder scene motion. 
For BEDLAM2.0, we split the tracking subset into 3,151 sequences for training and 300 sequences for testing. 
For DROID, we focus on wrist-camera videos and report video depth only, as wrist-camera pose annotations can be noisy for camera evaluation.
For fair and efficient evaluation across all new-domain benchmarks, we uniformly subsample each evaluation video by taking every 10th frame.
Regarding original-domain retention, we use established benchmarks from prior work~\cite{wang2025vggt, wang2024dust3r, wang2025pi, wang2025continuous}:  Sintel~\cite{bozic2021transformerfusion}, KITTI~\cite{geiger2013vision}, and Bonn~\cite{palazzolo2019refusion} for video depth, and RealEstate10K~\cite{zhou2018stereo} for camera estimation.

\vspace{0.5em}
\noindent \textbf{Evaluation Metrics.}
We report results for both video depth estimation and camera estimation. For video depth estimation, following prior work~\cite{wang2025continuous, zhang2024monst3r, wang2025pi}, we report Absolute Relative Error (Abs Rel) and threshold accuracy at $\delta < 1.25$ under two alignment settings: (i) scale-only alignment and (ii) joint scale and 3D translation alignment.
For camera estimation, we follow the angular-accuracy protocol in prior work~\cite{wang2025vggt, wang2024dust3r, wang2025pi}, and compute the Relative Rotation Accuracy (RRA) and Relative Translation Accuracy (RTA) as the percentage of pairs whose rotation/translation angular errors are below a threshold. 
We then report the Area Under the Curve (AUC) of the $\min(\mathrm{RRA}, \mathrm{RTA})$--threshold curve at thresholds of $5^\circ$, $15^\circ$, and $30^\circ$.

\begin{table}[t]
\centering
\scriptsize
\setlength{\tabcolsep}{2.8pt}
\renewcommand{\arraystretch}{1.10}
\caption{New-domain video depth evaluation on OmniGeo and OmniVideo~\cite{zhou2025omniworld}. 
}
\label{tab:vggt_new_domain_depth}
\resizebox{0.99\columnwidth}{!}{%
\begin{tabular}{lcccccccc}
\toprule
\multirow{3}{*}{Methods}
& \multicolumn{4}{c}{OmniGeo}
& \multicolumn{4}{c}{OmniVideo} \\
\cmidrule(lr){2-5}\cmidrule(lr){6-9}
& \multicolumn{2}{c}{scale}
& \multicolumn{2}{c}{scale\&shift}
& \multicolumn{2}{c}{scale}
& \multicolumn{2}{c}{scale\&shift} \\
\cmidrule(lr){2-3}\cmidrule(lr){4-5}
\cmidrule(lr){6-7}\cmidrule(lr){8-9}
& Abs Rel $\downarrow$ & $\delta<1.25$ $\uparrow$
& Abs Rel $\downarrow$ & $\delta<1.25$ $\uparrow$
& Abs Rel $\downarrow$ & $\delta<1.25$ $\uparrow$
& Abs Rel $\downarrow$ & $\delta<1.25$ $\uparrow$ \\
\midrule
VGGT~\cite{wang2025vggt}
& 0.346 & 0.592 & 0.180 & 0.758
& 0.236 & 0.713 & 0.145 & 0.824 \\
\ours(VGGT)
&\textbf{ 0.278} &\textbf{ 0.703 }& \textbf{0.124} & \textbf{0.867}
& \textbf{0.181} & \textbf{0.792} &\textbf{ 0.113} & \textbf{0.876} \\
\bottomrule
\end{tabular}%
}
\end{table}

\subsection{Main Results: Continual Self-improvement Evaluation}
\label{sec:cl_eval}
We evaluate continual self-improvement from two perspectives: \emph{new-domain adaptation} (improvement on the target domain) and \emph{original-domain retention} (preserving performance on original domains).
In this section, we focus on our primary setting: self-improving VGGT on OmniWorld-Game.

\vspace{0.25em}
\noindent \textbf{New-domain adaptation.}
As shown in Tab.~\ref{tab:vggt_new_domain_depth} and Tab.~\ref{tab:camera_overall}, self-improvement yields substantial gains over the pretrained baseline on Game-Benchmark for both video depth and camera estimation. This shows that our self improvement framework can effectively improve VGGT's performance on the target domain without accessing any ground-truth geometric supervision.

\vspace{0.25em}
\noindent \textbf{Original-domain retention.}
Crucially, these in-domain gains do not come at the expense of prior capabilities. Tab.~\ref{tab:camera_overall} and Tab.~\ref{tab:videodepth} show that after self-improvement, VGGT maintains, and in many cases improves, camera and video-depth estimation accuracy on standard evaluation benchmarks. This suggests that our framework does not overfit to the new domain. Instead, it improves the model’s capabilities more broadly while preserving previously acquired geometric priors.

\begin{table}[t]
\centering
\small
\setlength{\tabcolsep}{3.6pt}
\renewcommand{\arraystretch}{1.12}
\caption{Camera estimation: new-domain results are reported on OmniGeo and OmniVideo~\cite{zhou2025omniworld}, and original-domain retention is evaluated on RealEstate10K~\cite{zhou2018stereo}.
}
\label{tab:camera_overall}
\resizebox{0.98\columnwidth}{!}{%
\begin{tabular}{lccccccccc}
\toprule
\multirow{2}{*}{Methods}
& \multicolumn{3}{c}{OmniGeo}
& \multicolumn{3}{c}{OmniVideo}
& \multicolumn{3}{c}{RealEstate10K} \\
\cmidrule(lr){2-4}\cmidrule(lr){5-7}\cmidrule(lr){8-10}
& AUC@5 $\uparrow$ & AUC@15 $\uparrow$ & AUC@30 $\uparrow$
& AUC@5 $\uparrow$ & AUC@15 $\uparrow$ & AUC@30 $\uparrow$
& AUC@5 $\uparrow$ & AUC@15 $\uparrow$ & AUC@30 $\uparrow$ \\
\midrule
VGGT  & 45.093 & 64.659 & 72.649 & 67.785 & 83.468 & 89.743 & 38.597 & 66.404 & 78.833 \\
\ours(VGGT)  & \textbf{58.271} & \textbf{79.034} & \textbf{87.285} & \textbf{75.420} & \textbf{89.132} & \textbf{93.945} & \textbf{48.765} & \textbf{73.324} & \textbf{83.359} \\
\bottomrule
\end{tabular}%
}
\end{table}

\begin{table*}[t]
    \centering
    \caption{
        {Video Depth Estimation on original domains: Sintel~\cite{bozic2021transformerfusion}, Bonn~\cite{palazzolo2019refusion} and KITTI~\cite{geiger2013vision}.}
    }
    \scriptsize
    \setlength{\tabcolsep}{2.8pt}
    \renewcommand{\arraystretch}{0.95}

    \resizebox{\textwidth}{!}{
    \tablestyle{8.5pt}{1}
    \begin{tabular}{lcllllll}
        \toprule
        \multirow{2}{*}{\textbf{Method}} &
        \multirow{2}{*}{\textbf{Align}} &
        \multicolumn{2}{c}{\textbf{Sintel}} &
        \multicolumn{2}{c}{\textbf{Bonn}} &
        \multicolumn{2}{c}{\textbf{KITTI}} \\
        \cmidrule(r){3-4} \cmidrule(r){5-6} \cmidrule(r){7-8}
        & &
        Abs Rel $\downarrow$ & $\delta<1.25$ $\uparrow$ &
        Abs Rel $\downarrow$ & $\delta<1.25$ $\uparrow$ &
        Abs Rel $\downarrow$ & $\delta<1.25$ $\uparrow$ \\
        \midrule
        VGGT~\cite{wang2025vggt} & \multirow{2}{*}{scale}
        & 0.300 & 0.641 & 0.055 & 0.971 & 0.074 & 0.960 \\
        \textbf{\ours(VGGT)} & 
        & \textbf{0.283} & \textbf{0.707} & \textbf{0.046} & \textbf{0.977} & \textbf{0.047} & \textbf{0.981} \\
        
        \midrule
        VGGT~\cite{wang2025vggt} & \multirow{2}{*}{\makecell[c]{scale\& \\ shift}}
        & 0.227 & 0.684 & 0.049 & 0.974 & 0.059 & 0.961 \\
        \textbf{\ours(VGGT)} &
        & \textbf{0.212} & \textbf{0.692} & \textbf{0.044} & \textbf{0.977} & \textbf{0.042} & \textbf{0.979} \\
        \bottomrule
    \end{tabular}
    }
    \label{tab:videodepth}
\end{table*}

\begin{table}[t]
\centering
\small
\setlength{\tabcolsep}{3.8pt}
\renewcommand{\arraystretch}{1.08}
\caption{\textbf{Generality across base models and training data.}
We apply our self-improvement framework to VGGT and $\pi^3$ on two training sources without using annotation.
\textbf{Left:} self-improve on DROID and evaluate \emph{video depth} on DROID.
\textbf{Right:} self-improve on BEDLAM2.0 and evaluate \emph{video depth} and \emph{camera} on BEDLAM2.0.}
\label{tab:cl_generality_droid_bedlam2_compact}

\resizebox{0.98\columnwidth}{!}{%
\begin{tabular}{lcccc!{\vrule width 0.5pt}ccc!{\vrule width 0.5pt}cccc}
\toprule
\multirow{4}{*}{\textbf{Model}} &
\multicolumn{4}{c!{\vrule width 0.5pt}}{\textbf{Train/Eval: DROID}} &
\multicolumn{7}{c}{\textbf{Train/Eval: BEDLAM2.0}} \\
\cmidrule(lr){2-5}\cmidrule(lr){6-12}

& \multicolumn{4}{c!{\vrule width 0.5pt}}{\textbf{Video Depth}} &
\multicolumn{3}{c!{\vrule width 0.5pt}}{\textbf{Camera}} &
\multicolumn{4}{c}{\textbf{Video Depth}} \\
\cmidrule(lr){2-5}\cmidrule(lr){6-8}\cmidrule(lr){9-12}

& \multicolumn{2}{c}{\textbf{scale}} & \multicolumn{2}{c!{\vrule width 0.5pt}}{\textbf{scale\&shift}}
& \multirow{2}{*}{AUC@5$\uparrow$} & \multirow{2}{*}{AUC@15$\uparrow$} & \multirow{2}{*}{AUC@30$\uparrow$}
& \multicolumn{2}{c}{\textbf{scale}} & \multicolumn{2}{c}{\textbf{scale\&shift}} \\
\cmidrule(lr){2-3}\cmidrule(lr){4-5}\cmidrule(lr){9-10}\cmidrule(lr){11-12}

& Abs Rel$\downarrow$ & $\delta<1.25\uparrow$ & Abs Rel$\downarrow$ & $\delta<1.25\uparrow$
& & &
& Abs Rel$\downarrow$ & $\delta<1.25\uparrow$ & Abs Rel$\downarrow$ & $\delta<1.25\uparrow$ \\
\midrule
VGGT         & 0.306 & 0.597 & 0.294 & 0.645 & 77.22 & 91.79 & 95.84 & 0.110 & 0.921 & 0.086 & 0.934 \\
\ours(VGGT)  & \textbf{0.254} & \textbf{0.656} & \textbf{0.223} & \textbf{0.733} & \textbf{86.48} & \textbf{95.38} & \textbf{97.68} & \textbf{0.072} & \textbf{0.954} & \textbf{0.051} & \textbf{0.959} \\
\midrule
$\pi^3$        & 0.271 & 0.670 & 0.252 & 0.709 & 77.58 & 92.19 & 96.08 & 0.034 & 0.982 & 0.028 & 0.981 \\
\ours($\pi^3$) & \textbf{0.249} & \textbf{0.699} & \textbf{0.234} & \textbf{0.732} & \textbf{81.06} & \textbf{93.47} & \textbf{96.72} & \textbf{0.032} & \textbf{0.983} & \textbf{0.027} & \textbf{0.982} \\
\bottomrule
\end{tabular}%
}
\end{table}

\begin{figure}[t]
    \centering
    \includegraphics[width=\textwidth]{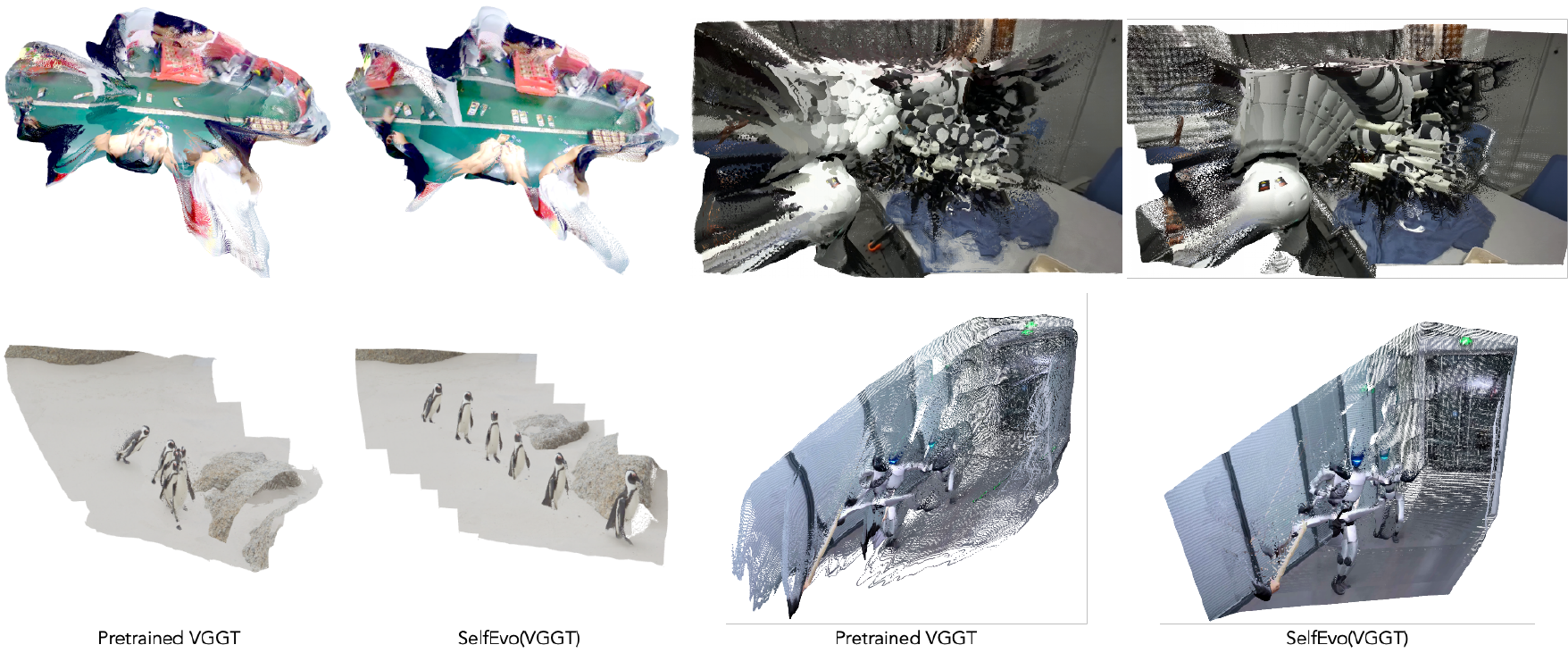}
    \caption{Visual result on unseen-domain data, including animal motion, robotics and ego-centric.
    }
    \label{fig:ood}
\end{figure}

\subsection{Generality Across Base Models and Training Data}
\label{sec:cl_generality}
We run additional experiments to test generality across base models and unlabeled training sources. These experiments verify that the gains are not tied to a specific model or dataset, but reflect a broadly applicable self-improvement mechanism across architectures and data regimes.

\vspace{0.5em}
\noindent \textbf{$\pi^3$ with BEDLAM2.0 and DROID.}
We  apply our framework to $\pi^3$~\cite{wang2025pi} on BEDLAM2.0. Tab.~\ref{tab:cl_generality_droid_bedlam2_compact} shows that video-depth performance is comparable to the pretrained model, while camera estimation improves substantially; the small depth change likely reflects BEDLAM2.0’s relatively simple geometry and strong pretrained depth. We also train $\pi^3$ on DROID and evaluate on the wrist-camera subset. We observe consistent gains in video depth (Tab.~\ref{tab:cl_generality_droid_bedlam2_compact}). Due to noisy wrist-camera pose annotations, we do not report camera metrics on DROID.

\vspace{0.5em}
\noindent \textbf{VGGT with BEDLAM2.0 and DROID.}
We also apply the same framework to VGGT on BEDLAM2.0 and DROID. As shown in Tab.~\ref{tab:cl_generality_droid_bedlam2_compact}, VGGT similarly benefits from self-improvement on these datasets, further supporting that our framework generalizes across both model architectures and data regimes.

\begin{figure}[t]
    \centering
    \includegraphics[width=\textwidth]{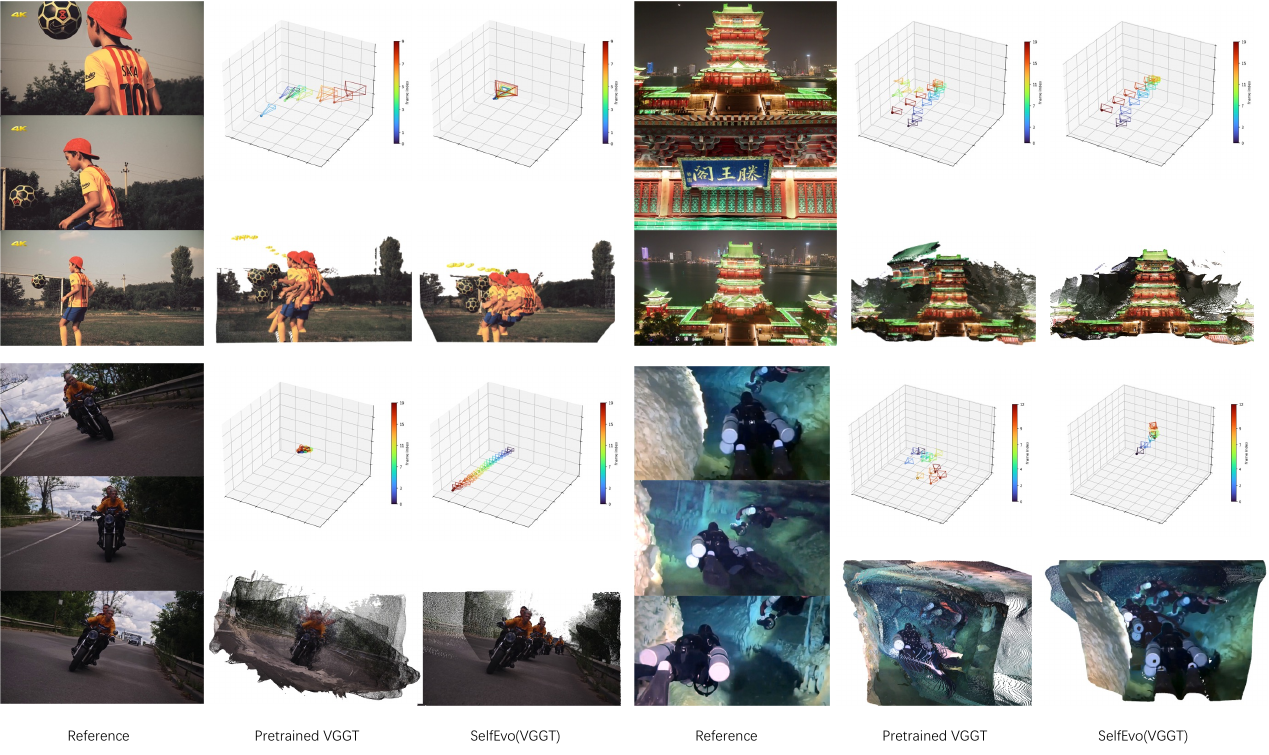}
    \caption{Qualitative results for camera and geometry on in the wild videos.}
    \label{fig:qualitative4}
\end{figure}

\begin{table}[t]
\centering
\scriptsize
\setlength{\tabcolsep}{2.8pt}
\renewcommand{\arraystretch}{1.10}
\caption{\textbf{unseen domain evaluation.} unseen domain evaluation for VGGT self-improved on OmniWorld-Game~\cite{zhou2025omniworld} and then evaluate on DROID~\cite{khazatsky2024droid} and HOI4D~\cite{Liu_2022_CVPR}. 
}
\label{tab:vggt_ood_video_depth}
\resizebox{0.99\columnwidth}{!}{%
\begin{tabular}{lcccccccc}
\toprule
\multirow{3}{*}{Methods}
& \multicolumn{4}{c}{DROID}
& \multicolumn{4}{c}{HOI4D} \\
\cmidrule(lr){2-5}\cmidrule(lr){6-9}
& \multicolumn{2}{c}{scale}
& \multicolumn{2}{c}{scale\&shift}
& \multicolumn{2}{c}{scale}
& \multicolumn{2}{c}{scale\&shift} \\
\cmidrule(lr){2-3}\cmidrule(lr){4-5}
\cmidrule(lr){6-7}\cmidrule(lr){8-9}
& Abs Rel $\downarrow$ & $\delta<1.25$ $\uparrow$
& Abs Rel $\downarrow$ & $\delta<1.25$ $\uparrow$
& Abs Rel $\downarrow$ & $\delta<1.25$ $\uparrow$
& Abs Rel $\downarrow$ & $\delta<1.25$ $\uparrow$ \\
\midrule
VGGT~\cite{wang2025vggt}
& 0.306 & 0.597 & 0.294 & 0.645
& 0.044 & 0.957 & 0.041 & 0.966 \\
\ours(VGGT)
& \textbf{0.268} & \textbf{0.659 }& \textbf{0.237} & \textbf{0.724}
& \textbf{0.030} & \textbf{0.988} & \textbf{0.031} & \textbf{0.987} \\
\bottomrule
\end{tabular}%
0.30}
\end{table}

\subsection{Unseen-domain Generalization}
\label{sec:ood_generalization}
Our previous evaluation measures in-distribution adaptation and original-domain retention, but it does not directly answer whether the gains obtained from self-improvement transfer \emph{beyond} the adaptation domain. We therefore conduct an additional unseen-domain generalization study under the main setting. For the unseen-domain setting, we refer to dataset domains on which the pretrained model was neither trained nor evaluated. We self-improve \textsc{VGGT} on OmniWorld-Game~\cite{zhou2025omniworld}, but test on DROID~\cite{khazatsky2024droid}, a robot manipulation dataset, and HOI4D~\cite{Liu_2022_CVPR}, an egocentric video dataset. As shown in Tab.~\ref{tab:vggt_ood_video_depth}, self-improvement consistently outperforms the pretrained baseline, indicating that the gains transfer beyond the adaptation domain. We additionally provide qualitative comparisons on diverse in-the-wild videos spanning {animal motion}, {egocentric} videos and {robotics} scenarios in Fig.~\ref{fig:ood} and Fig.~\ref{fig:qualitative4}.

\section{Analysis}
\label{sec:analysis}
In this section, we conduct a series of experiments to understand why such a simple framework can yield significant improvements in performance. {Unless stated otherwise, all analysis experiments follow our primary setting: VGGT self-improvement on OmniWorld-Game~\cite{zhou2025omniworld}.}

\begin{table}[t]
\centering
\small
\setlength{\tabcolsep}{4pt}
\renewcommand{\arraystretch}{1.15}

\caption{\textbf{Ablation of asymmetry mechanisms on VGGT.}
``aug-stu'' augments student inputs only, while ``aug-all'' augments both the teacher and student inputs.
}
\label{tab:ablation_asymmetry_omnigeo_omnivideo}
\resizebox{1.0 \columnwidth}{!}{
\begin{tabular}{llcccccc}
\toprule
\multirow{3}{*}{Dataset} & \multirow{3}{*}{Methods}
& \multicolumn{4}{c}{Video Depth} & \multicolumn{2}{c}{Camera} \\
\cmidrule(lr){3-6}\cmidrule(lr){7-8}
& & \multicolumn{2}{c}{scale} & \multicolumn{2}{c}{scale\&shift}
& \multirow{2}{*}{AUC@15$\uparrow$} & \multirow{2}{*}{AUC@30$\uparrow$} \\
\cmidrule(lr){3-4}\cmidrule(lr){5-6}
& & Abs Rel$\downarrow$ & $\delta<1.25\uparrow$ & Abs Rel$\downarrow$ & $\delta<1.25\uparrow$ & & \\
\midrule

\multirow{4}{*}{OmniGeo}
& aug-stu    & 0.347 & 0.619 & 0.148 & 0.824 & 66.476 & 75.675 \\
& aug-all    & 0.364 & 0.607 & 0.150 & 0.820 & 64.432 & 73.890 \\
& cropping   & 0.384 & 0.605 & 0.147 & 0.829 & 64.262 & 75.990 \\
& dropping   & \textbf{0.296} &\textbf{ 0.682} &\textbf{ 0.127} & \textbf{0.862} & \textbf{75.317} & \textbf{84.323} \\

\bottomrule
\end{tabular}
}
\end{table}

\begin{table}[t]
\centering
\small
\setlength{\tabcolsep}{4pt}
\renewcommand{\arraystretch}{1.15}

\caption{\textbf{Ablation of frame selection strategy on VGGT.}
}
\label{tab:ablation_frame_sample}
\resizebox{1.0 \columnwidth}{!}{
\begin{tabular}{llcccccc}
\toprule
\multirow{3}{*}{Dataset} & \multirow{3}{*}{Methods}
& \multicolumn{4}{c}{Video Depth} & \multicolumn{2}{c}{Camera} \\
\cmidrule(lr){3-6}\cmidrule(lr){7-8}
& & \multicolumn{2}{c}{scale} & \multicolumn{2}{c}{scale\&shift}
& \multirow{2}{*}{AUC@15$\uparrow$} & \multirow{2}{*}{AUC@30$\uparrow$} \\
\cmidrule(lr){3-4}\cmidrule(lr){5-6}
& & Abs Rel$\downarrow$ & $\delta<1.25\uparrow$ & Abs Rel$\downarrow$ & $\delta<1.25\uparrow$ & & \\
\midrule

\multirow{3}{*}{OmniGeo}
& keep-top    & 0.347 & 0.619 & 0.148 & 0.824 & 71.423 & 80.903 \\
& keep-bottom    & 0.364 & 0.607 & 0.150 & 0.820 & 78.521 & 87.013 \\
& random   &\textbf{ 0.278} &\textbf{ 0.703 }& \textbf{0.124} & \textbf{0.867} & \textbf{{79.034}} & \textbf{{87.285}} \\

\bottomrule
\end{tabular}
}
\end{table}

\begin{table}[t]
\centering
\small
\setlength{\tabcolsep}{4pt}
\renewcommand{\arraystretch}{1.15}

\caption{\textbf{Ablation of online supervision.}
We self-improve on OmniWorld-Game~\cite{zhou2025omniworld} and evaluate on its benchmark OmniGeo.
}
\label{tab:ablation_online}
\resizebox{1.0 \columnwidth}{!}{
\begin{tabular}{llcccccc}
\toprule
\multirow{3}{*}{Dataset} & \multirow{3}{*}{Mode}
& \multicolumn{4}{c}{Video Depth} & \multicolumn{2}{c}{Camera} \\
\cmidrule(lr){3-6}\cmidrule(lr){7-8}
& & \multicolumn{2}{c}{scale} & \multicolumn{2}{c}{scale\&shift}
& \multirow{2}{*}{AUC@15$\uparrow$} & \multirow{2}{*}{AUC@30$\uparrow$} \\
\cmidrule(lr){3-4}\cmidrule(lr){5-6}
& & Abs Rel$\downarrow$ & $\delta<1.25\uparrow$ & Abs Rel$\downarrow$ & $\delta<1.25\uparrow$ & & \\
\midrule

\multirow{2}{*}{OmniGeo}
& offline    & 0.357 & 0.615 & 0.150 & 0.817 & 64.618 & 74.939 \\
& online   &\textbf{ 0.278} &\textbf{ 0.703 }& \textbf{0.124} & \textbf{0.867} & \textbf{79.034} & \textbf{87.285} \\

\bottomrule
\end{tabular}
}
\end{table}

\begin{table}[t]
\centering
\small
\setlength{\tabcolsep}{3.6pt}
\renewcommand{\arraystretch}{1.15}
\caption{
\textbf{Role of Online Supervision in OOD transfer after target domain adaptation.}
We compare {Pretrain} (baseline), {SFT} ({Offline}, fully supervised fine-tuning on the target domain) and {\ours} ({Online}, unlabeled self-improvement).
}
\label{tab:self-vs-SFT}
\resizebox{0.98\columnwidth}{!}{%
\begin{tabular}{lcccccccc}
\toprule
\multirow{3}{*}{Mode} 
& \multicolumn{4}{c}{Droid~\cite{khazatsky2024droid}} 
& \multicolumn{4}{c}{Bonn~\cite{palazzolo2019refusion}} \\
\cmidrule(lr){2-5}\cmidrule(lr){6-9}
& \multicolumn{2}{c}{scale} & \multicolumn{2}{c}{scale \& shift}
& \multicolumn{2}{c}{scale} & \multicolumn{2}{c}{scale \& shift} \\
\cmidrule(lr){2-3}\cmidrule(lr){4-5}\cmidrule(lr){6-7}\cmidrule(lr){8-9}
& Abs Rel $\downarrow$ & $\delta<1.25$ $\uparrow$
& Abs Rel $\downarrow$ & $\delta<1.25$ $\uparrow$
& Abs Rel $\downarrow$ & $\delta<1.25$ $\uparrow$
& Abs Rel $\downarrow$ & $\delta<1.25$ $\uparrow$ \\
\midrule
Pretrain (baseline)
& 0.306 & 0.597
& 0.294 & 0.645
& 0.055 & 0.971
& 0.049 & 0.974 \\
SFT (Offline)
& \textbf{0.281} & \textbf{0.658}
& 0.264 & 0.692
& 0.081 & 0.895
& 0.073 & 0.940 \\
\ours (Online)
& {${0.285}$} & 0.622
& {$\mathbf{0.243}$} & \textbf{0.703}
& {$\mathbf{0.049}$} & \textbf{0.973}
& {$\mathbf{0.044}$}  &\textbf{ 0.976} \\
\bottomrule
\end{tabular}%
}
\end{table}

\subsection{What Makes an Effective Teacher--Student Asymmetry?}
\label{sec:ablation_asymmetry}
We study how the {form} of context asymmetry affects self improvement, comparing appearance perturbations (color jitter, grayscale), crop-based asymmetry, and frame-dropping asymmetry.
For appearance-level asymmetry, we test two variants: (i) applying augmentations only to the student while the teacher sees original frames, and (ii) applying augmentations to both teacher and student with independently sampled parameters from the same distribution~\cite{oquab2023dinov2}. For crop-based asymmetry, the teacher observes full frames while the student receives cropped views.
For a fair comparison, we keep other settings fixed and use a smaller batch size for efficiency; the batch size is matched to the number of student frames which contribute gradients, not teacher frames.
As shown in Tab.~\ref{tab:ablation_asymmetry_omnigeo_omnivideo}, frame dropping performs best. 
This suggests that self-improvement benefits from effective spatiotemporal context asymmetry.

\subsection{Frame Selection Strategy}
\label{sec:frame_selection}

We study a core design choice about how to select the student frames from the teacher’s longer sequence in this section. For each clip, the teacher receives $m\!\in\![m_l,m_h]$ frames; we sample a shorter subsequence of length $n\!\in\![n_l,n_h]$ for the student and supervise it with the teacher’s predictions on the same frames. 

\vspace{0.5em}
\noindent \textbf{Random sampling.}
Given a teacher sequence of length $m$, we first sample a target student length $n\in[n_l,n_h]$.
We then draw $n$ frame indices at random.

\vspace{0.5em}
\noindent \textbf{Attention-guided sampling.}
Motivated by~\cite{han2025emergent}, we also explored selecting frames based on per-frame \emph{importance} scores extracted from the teacher during inference. Concretely, we compute a frame-to-frame attention matrix from a chosen transformer layer by using patch tokens of each frame as queries and all frames as keys/values, and then reduce it to a single score per frame by aggregating the \emph{incoming} attention mass (averaged across heads and query patches, excluding self-attention). To avoid an over-dominant reference effect, we construct this attention matrix after \emph{removing the first frame} from both the query set and the key/value set. We min--max normalize the resulting per-frame scores within each sequence. In addition, we optionally compute a {feature-map score} by taking per-frame aggregated tokens and measuring their cosine-based affinity, and mix it with the attention score to form a more stable signal. Given these scores, we tried two selection rules: \textbf{keep-top} (prefer high-score frames) and \textbf{keep-bottom} (prefer low-score frames). 

We report the results in Tab~\ref{tab:ablation_frame_sample}. Although attention-guided strategies can be reasonable heuristics, we found that  random sampling 
is the most robust and consistently yields the best performance across models and benchmarks.

\begin{table}[t]
\centering
\small
\setlength{\tabcolsep}{4pt}
\renewcommand{\arraystretch}{1.15}

\caption{\textbf{Ablation of training recipe on VGGT.}
{freeze-A/C/D} denotes freezing the aggregator, camera head, or depth head, respectively; {freeze-C\&D} freezes both heads; {train-all} updates all modules. Our default is {freeze-C (ours)}.}
\label{tab:ablation_training_recipe}
\resizebox{1.0 \columnwidth}{!}{
\begin{tabular}{llcccccc}
\toprule
\multirow{3}{*}{Dataset} & \multirow{3}{*}{Methods}
& \multicolumn{4}{c}{Video Depth} & \multicolumn{2}{c}{Camera} \\
\cmidrule(lr){3-6}\cmidrule(lr){7-8}
& & \multicolumn{2}{c}{scale} & \multicolumn{2}{c}{scale\&shift}
& \multirow{2}{*}{AUC@15$\uparrow$} & \multirow{2}{*}{AUC@30$\uparrow$} \\
\cmidrule(lr){3-4}\cmidrule(lr){5-6}
& & Abs Rel$\downarrow$ & $\delta<1.25\uparrow$ & Abs Rel$\downarrow$ & $\delta<1.25\uparrow$ & & \\
\midrule

\multirow{5}{*}{OmniGeo}
& freeze-D    & 0.325 & 0.654 & 0.134 & 0.851 & 77.982 & 86.540 \\
& freeze-A    & 0.370 & 0.613 & 0.162 & 0.793 & 62.270 & 71.497 \\
& freeze-C\&D   & 0.307 & 0.669 & 0.133 & 0.855 & 77.511 & 86.236 \\
& train-all   & 0.292 & 0.696 & 0.125 & 0.865 & 78.704 & 86.807 \\
& freeze-C(ours)   & \textbf{ 0.278} &\textbf{ 0.703 }& \textbf{0.124} & \textbf{0.867} & \textbf{79.034} & \textbf{87.285} \\

\bottomrule
\end{tabular}
}
\end{table}

\subsection{The Role of Online Supervision}
\label{sec:online_vs_offline}

We next study the role of \emph{online} teacher updating in our self-improvement setting by comparing it against an \emph{offline} pseudo-label fine-tuning baseline (Tab.~\ref{tab:ablation_online}). In the offline setting, pseudo targets are generated once by a fixed pretrained model and then used to supervise subsequent training. In the online setting, the teacher is an EMA-smoothed copy of the student and is updated throughout training, allowing the pseudo targets to evolve together with the student.

This difference is critical for continual adaptation. With a fixed offline teacher, the supervision quickly becomes stale as the student drifts from the initial pretrained distribution, which can amplify early pseudo-label errors and lead to unstable specialization. In contrast, an online EMA teacher co-evolves with the student, providing a more consistent target that tracks training dynamics and yields progressively stronger supervision. 

Finally, we compare self-improvement not only to pseudo-label fine-tuning, but also to \emph{fully supervised} fine-tuning on the target domain (Tab.~\ref{tab:self-vs-SFT}).
While supervised fine-tuning (SFT) can improve {target-domain} performance, its behavior is closer to aggressive domain-specific adaptation:
the model is encouraged to fit target-domain appearance, motion, and camera statistics, which can lead to {reduced transferability} to substantially different domains (e.g., Bonn and DROID), and can underperform our {online} annotation-free self-improvement on OOD benchmarks as shown in the Tab.~\ref{tab:self-vs-SFT}.
By comparison, our online framework are constrained by consistency and EMA smoothing, which mitigates representation drift and better preserves cross-domain geometric priors.

\subsection{Training Recipe Choices: What to Update?}
\label{sec:what_to_update}
A practical question in continual post-training is which components of a pretrained multi-view reconstruction model should be updated. Since pseudo supervision is imperfect and can shift under self-distillation context asymmetry, we compare freezing the camera decoder, depth decoder, and shared backbone (and combinations) with all other settings fixed.
As shown in Tab.~\ref{tab:ablation_training_recipe}, freezing \emph{only} the camera decoder yields the best overall performance: it provides the most consistent gains on both video-depth and camera metrics.

We attribute this to frame dropping, which changes the spatiotemporal context seen by the student. Because camera estimation relies on global, sequence-level cues aggregated across frames, mismatched teacher--student frame subsets can distort camera pseudo-supervision. Freezing the camera decoder anchors camera prediction to the pretrained solution while allowing the backbone and depth decoder to improve, yielding a better stability--plasticity trade-off.

\begin{table}[t]
\centering
\small
\setlength{\tabcolsep}{4pt}
\renewcommand{\arraystretch}{1.15}

\caption{\textbf{Ablation of supervision form.}
 $+\mathcal{L}_{\text{feat}}$ add feature matching loss. 
}
\label{tab:ablation_loss}
\resizebox{1.0 \columnwidth}{!}{
\begin{tabular}{llcccccc}
\toprule
\multirow{3}{*}{Dataset} & \multirow{3}{*}{Mode}
& \multicolumn{4}{c}{Video Depth} & \multicolumn{2}{c}{Camera} \\
\cmidrule(lr){3-6}\cmidrule(lr){7-8}
& & \multicolumn{2}{c}{scale} & \multicolumn{2}{c}{scale\&shift}
& \multirow{2}{*}{AUC@15$\uparrow$} & \multirow{2}{*}{AUC@30$\uparrow$} \\
\cmidrule(lr){3-4}\cmidrule(lr){5-6}
& & Abs Rel$\downarrow$ & $\delta<1.25\uparrow$ & Abs Rel$\downarrow$ & $\delta<1.25\uparrow$ & & \\
\midrule

\multirow{2}{*}{OmniGeo}
& $+\mathcal{L}_{\text{feat}}$  & 0.283 & 0.696 & 0.125 & \textbf{0.867} & 78.935 & \textbf{87.310} \\
& base   &\textbf{ 0.278} &\textbf{ 0.703 }& \textbf{0.124} & \textbf{0.867} & \textbf{79.034} & {87.285} \\

\bottomrule
\end{tabular}
}
\end{table}

\subsection{Supervision form: output distillation vs.\ feature matching.}
\label{sec:loss_design}
Beyond output-level distillation, we optionally add an intermediate feature-matching loss $\mathcal{L}_{\text{feat}}$ to stabilize learning under severe teacher--student asymmetry.
Concretely, we extract per-frame representations from the teacher's aggregator tokens at multiple layers (e.g., $\{4,11,17,23\}$) by mean-pooling patch tokens into a single feature vector per frame, and match teacher and student features on the selected student frames.
Overall, we find that $\mathcal{L}_{\text{feat}}$ brings \emph{no consistent improvement}: performance is largely unchanged with or without feature matching across datasets (Tab.~\ref{tab:ablation_loss}).
This suggests that output-level supervision already provides a sufficiently strong learning signal for self-improvement, while feature alignment under  spatiotemporal context asymmetry is at best redundant.

\section{Conclusion}
\textbf{Limitation.}
Our framework is most effective in settings with sufficient camera motion, where frame dropping provides a strong context asymmetry signal. When the camera remains static, it is difficult to create asymmetry through frame dropping alone. Future work could extend frame-level selection to the token level by selectively dropping tokens for greater flexibility. Additionally, as with other self-improving frameworks, the absence of ground-truth supervision means extended training may risk model collapse. In practice, however, we generally observe stable performance without significant degradation. Understanding how to sustain improvement over longer training horizons remains an important direction for future work.

\vspace{0.5em}
\noindent\textbf{Conclusion}. This work systematically investigated how to continually improve pretrained multi-view reconstruction models {without} labeled 3D data, and identified key ingredients that made self-improvement effective and stable. Through extensive comparisons over design variants, we showed that an online self-distillation loop driven by spatiotemporal context asymmetry (with per-step EMA updates) provided reliable training signals for geometry prediction, especially in dynamic scenes. We further found that simple choices such as random frame dropping for inducing asymmetry, using output-level loss, and selectively freezing the camera decoder were crucial for robust gains. Across diverse benchmarks spanning multiple domains and base models, \ours consistently improved pretrained baselines while largely preserving performance on established evaluation domains.

\bibliographystyle{splncs04}
\bibliography{main}
\clearpage
\setcounter{page}{1}
\appendix

\begin{center}
    {\LARGE \bfseries Self-Improving 4D Perception via Self-Distillation}
\end{center}

\begin{center}
    {Supplementary Material}
\end{center}

\paragraph{Overview.}
This supplementary material is organized as follows. In Sec.~\ref{sec:supp_pre}, we provide additional details for the preliminary context analysis and further show how increasing temporal context improves feedforward reconstruction quality. In Sec.~\ref{sec:supp_impl}, we present additional implementation details of the self-improving procedure. In Sec.~\ref{sec:supp_analysis_omnivideo}, we report additional ablation results on OmniVideo~\cite{zhou2025omniworld}, including teacher--student asymmetry, frame selection strategy, online versus offline supervision, training recipe choices, and supervision form. In Sec.~\ref{sec:supp_limitation}, we discuss an additional limitation of the proposed framework. In Sec.~\ref{sec:supp_continuous_self_improvement}, we provide checkpoint-wise evaluations on held-out OmniGeo and OmniVideo benchmarks, showing continuous self-improvement throughout training. In Sec.~\ref{sec:supp_additional_data}, we present qualitative results on additional training data sources, including Egocentric-10K~\cite{buildaiegocentric10k2025} and DynPose-100K~\cite{rockwell2025dynpose}. Finally, please visit our project page for better visualization: \url{https://self-evo.github.io/}.

\section{Details of the Preliminary Context Analysis}
\label{sec:supp_pre}

This section provides additional details for the preliminary context analysis discussed in Sec.~\ref{sec:pre} of the main paper. Our goal is to isolate how the amount of temporal context affects feedforward reconstruction quality.
We start from a low-context setting defined by two anchor frames sampled from the same video and separated by a large temporal gap. We then progressively increase the context by randomly sampling intermediate frames between the two anchors. For each context level, we run inference on the full set consisting of the two anchors and the sampled intermediate frames, while evaluating performance only on the two anchor frames. In this way, the added frames serve purely as context, allowing us to measure how additional observations affect the quality of the predictions on fixed targets.

\begin{figure}
    \centering
    \includegraphics[width=\linewidth]{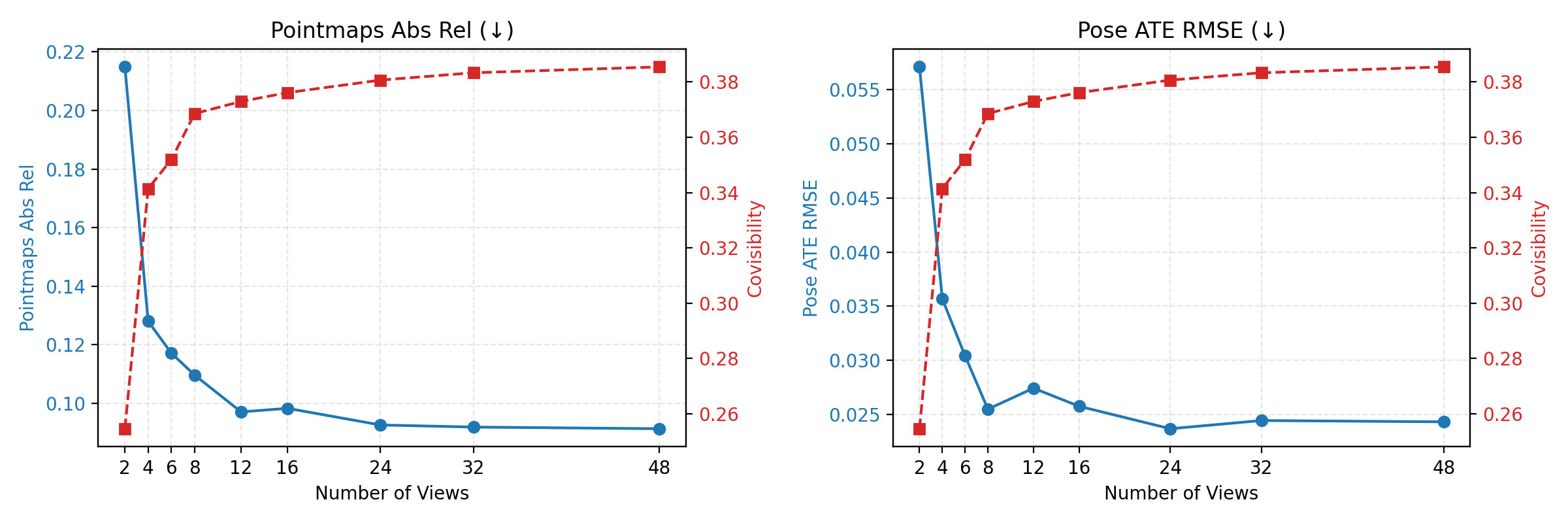}
    \caption{\small \textbf{Context improves feedforward reconstruction quality.}
    Starting from two temporally distant anchor frames, we progressively add intermediate frames as context and evaluate performance only on the anchors. As the number of input views increases, the overall covisibility increases, while both pointmap and pose errors decrease.}
    \label{fig:covis_analysis}
\end{figure}

We perform this analysis on ScanNet~\cite{dai2017scannet} and report both pointmap and camera pose errors as the context grows. In addition, we compute the overall covisibility score~\cite{keetha2025mapanything} to quantify the amount of shared scene content among the input views. The results are shown in Fig.~\ref{fig:covis_analysis}. As more intermediate views are introduced, the covisibility increases and the prediction errors decrease consistently. This trend supports the key assumption behind our framework: richer spatiotemporal context leads to stronger predictions, which can in turn be converted into a supervision signal through context asymmetry.

\section{Additional Implementation Details}
\label{sec:supp_impl}
Unless stated otherwise, during self-improving we freeze the camera decoder and update the remaining modules; the effect of this design choice is analyzed in Sec.~\ref{sec:what_to_update}. The student is trained for 20 epochs, each with 50 optimization steps. The teacher is updated after every optimization step using EMA with decay $\lambda$, where $\lambda=0.995$ for \textsc{VGGT} and $\lambda=0.99$ for \textsc{$\pi^3$}. We use a slightly smaller decay for \textsc{$\pi^3$} because its aggregator is deeper (36 layers versus 24 in \textsc{VGGT}), which empirically benefits from a more responsive teacher. For each training clip, the teacher input length is sampled as $m \in [m_l, m_h]$ with $m_l=24$ and $m_h=64$, while the student receives a subset of length $n \in [n_l, n_h]$ with $n_l=2$ and $n_h=12$. We use a composite learning-rate schedule with a 5\% linear warmup from $1\times10^{-8}$ to $1\times10^{-5}$, followed by a 95\% cosine decay from $1\times10^{-5}$ to $1\times10^{-8}$. All other training details follow the original pretraining recipe of the base model, except that ground-truth supervision is replaced by teacher-generated pseudo targets.

\begin{table}[t]
\centering
\small
\setlength{\tabcolsep}{4pt}
\renewcommand{\arraystretch}{1.15}
\caption{\textbf{OmniVideo counterpart of the asymmetry ablation on VGGT.}
``aug-stu'' augments student inputs only, while ``aug-all'' augments both the teacher and student inputs.}
\label{tab:supp_ablation_asymmetry_omnivideo}
\resizebox{1.0\columnwidth}{!}{
\begin{tabular}{llcccccc}
\toprule
\multirow{3}{*}{Dataset} & \multirow{3}{*}{Methods}
& \multicolumn{4}{c}{Video Depth} & \multicolumn{2}{c}{Camera} \\
\cmidrule(lr){3-6}\cmidrule(lr){7-8}
& & \multicolumn{2}{c}{scale} & \multicolumn{2}{c}{scale\&shift}
& \multirow{2}{*}{AUC@15$\uparrow$} & \multirow{2}{*}{AUC@30$\uparrow$} \\
\cmidrule(lr){3-4}\cmidrule(lr){5-6}
& & Abs Rel$\downarrow$ & $\delta<1.25\uparrow$ & Abs Rel$\downarrow$ & $\delta<1.25\uparrow$ & & \\
\midrule
\multirow{4}{*}{OmniVideo}
& aug-stu    & 0.211 & 0.759 & 0.125 & 0.857 & 86.602 & 92.303 \\
& aug-all    & 0.202 & 0.740 & 0.129 & 0.851 & 85.970 & 91.851 \\
& cropping   & 0.228 & 0.714 & 0.136 & 0.841 & 82.299 & 90.405 \\
& dropping   & \textbf{0.173} & \textbf{0.810} & \textbf{0.111} & \textbf{0.879} & \textbf{87.611} & \textbf{92.480} \\
\bottomrule
\end{tabular}
}
\end{table}

\begin{table}
\centering
\small
\setlength{\tabcolsep}{4pt}
\renewcommand{\arraystretch}{1.15}
\caption{\textbf{OmniVideo counterpart of the frame selection ablation on VGGT.}}
\label{tab:supp_ablation_frame_sample_omnivideo}
\resizebox{1.0\columnwidth}{!}{
\begin{tabular}{llcccccc}
\toprule
\multirow{3}{*}{Dataset} & \multirow{3}{*}{Methods}
& \multicolumn{4}{c}{Video Depth} & \multicolumn{2}{c}{Camera} \\
\cmidrule(lr){3-6}\cmidrule(lr){7-8}
& & \multicolumn{2}{c}{scale} & \multicolumn{2}{c}{scale\&shift}
& \multirow{2}{*}{AUC@15$\uparrow$} & \multirow{2}{*}{AUC@30$\uparrow$} \\
\cmidrule(lr){3-4}\cmidrule(lr){5-6}
& & Abs Rel$\downarrow$ & $\delta<1.25\uparrow$ & Abs Rel$\downarrow$ & $\delta<1.25\uparrow$ & & \\
\midrule
\multirow{4}{*}{OmniVideo}
& random         & \textbf{0.181} & \textbf{0.792} & \textbf{0.113} & \textbf{0.876} & \textbf{89.132} & \textbf{93.945} \\
& keep-top       & 0.329 & 0.632 & 0.133 & 0.853 & 86.602 & 92.303 \\
& keep-bottom    & 0.271 & 0.713 & 0.130 & 0.850 & 85.970 & 91.851 \\
& probabilistic  & 0.268 & 0.698 & 0.126 & 0.866 & 82.299 & 90.405 \\
\bottomrule
\end{tabular}
}
\end{table}
\begin{table}[t]
\centering
\small
\setlength{\tabcolsep}{4pt}
\renewcommand{\arraystretch}{1.15}
\caption{\textbf{OmniVideo counterpart of the online supervision ablation.}
We self-improve on OmniWorld-Game~\cite{zhou2025omniworld} and evaluate on its benchmark OmniVideo.}
\label{tab:supp_ablation_online_omnivideo}
\resizebox{1.0\columnwidth}{!}{
\begin{tabular}{llcccccc}
\toprule
\multirow{3}{*}{Dataset} & \multirow{3}{*}{Mode}
& \multicolumn{4}{c}{Video Depth} & \multicolumn{2}{c}{Camera} \\
\cmidrule(lr){3-6}\cmidrule(lr){7-8}
& & \multicolumn{2}{c}{scale} & \multicolumn{2}{c}{scale\&shift}
& \multirow{2}{*}{AUC@15$\uparrow$} & \multirow{2}{*}{AUC@30$\uparrow$} \\
\cmidrule(lr){3-4}\cmidrule(lr){5-6}
& & Abs Rel$\downarrow$ & $\delta<1.25\uparrow$ & Abs Rel$\downarrow$ & $\delta<1.25\uparrow$ & & \\
\midrule
\multirow{2}{*}{OmniVideo}
& offline    & 0.244 & 0.703 & 0.145 & 0.818 & 85.684 & 91.793 \\
& online     & \textbf{0.181} & \textbf{0.792} & \textbf{0.113} & \textbf{0.876} & \textbf{89.132} & \textbf{93.945} \\
\bottomrule
\end{tabular}
}
\end{table}

\section{Additional Analysis on OmniVideo}
\label{sec:supp_analysis_omnivideo}

Due to space constraints, Sec.~\ref{sec:analysis} in the main paper reports the ablation studies on OmniGeo only. We choose OmniGeo for the main-paper analysis because it is more directly aligned with our target task and serves as a stronger stress test of geometric prediction. Specifically, OmniGeo (\emph{3D Geometric Prediction Benchmark}) and OmniVideo (\emph{Camera-Controlled Video Generation Benchmark}) emphasize different aspects of evaluation. Compared with OmniVideo, OmniGeo typically involves more challenging geometry and therefore provides a more diagnostic benchmark for analysis. Here we provide the corresponding OmniVideo results for completeness. Overall, the conclusions remain consistent with those in the main paper.

\subsection{What Makes an Effective Teacher--Student Asymmetry?}
\label{sec:supp_ablation_asymmetry_omnivideo}

Tab.~\ref{tab:supp_ablation_asymmetry_omnivideo} shows the OmniVideo counterpart of Tab.~\ref{tab:ablation_asymmetry_omnigeo_omnivideo}. Consistent with the main paper, frame dropping performs best.

\subsection{Frame Selection Strategy}
\label{sec:supp_frame_selection_omnivideo}

Tab.~\ref{tab:supp_ablation_frame_sample_omnivideo} shows the OmniVideo counterpart of Tab.~\ref{tab:ablation_frame_sample}. Random sampling remains the strongest strategy.

\subsection{The Role of Online Supervision}
\label{sec:supp_online_vs_offline_omnivideo}

Tab.~\ref{tab:supp_ablation_online_omnivideo} shows the OmniVideo counterpart of Tab.~\ref{tab:ablation_online}. The same trend holds: online self-improvement substantially outperforms offline pseudo-label fine-tuning.

\subsection{Training Recipe Choices: What to Update?}
\label{sec:supp_what_to_update_omnivideo}

Tab.~\ref{tab:supp_ablation_training_recipe_omnivideo} shows the OmniVideo counterpart of Tab.~\ref{tab:ablation_training_recipe}. Freezing the camera decoder remains the best overall choice.

\subsection{Supervision Form: Output Distillation vs.\ Feature Matching}
\label{sec:supp_loss_design_omnivideo}

Tab.~\ref{tab:supp_ablation_loss_omnivideo} shows the OmniVideo counterpart of Tab.~\ref{tab:ablation_loss}. Adding feature matching does not provide consistent improvement.

\begin{table}
\centering
\small
\setlength{\tabcolsep}{4pt}
\renewcommand{\arraystretch}{1.15}
\caption{\textbf{OmniVideo counterpart of the training recipe ablation on VGGT.}
{freeze-A/C/D} denotes freezing the aggregator, camera head, or depth head, respectively; {freeze-C\&D} freezes both heads; {train-all} updates all modules. Our default is {freeze-C (ours)}.}
\label{tab:supp_ablation_training_recipe_omnivideo}
\resizebox{1.0\columnwidth}{!}{
\begin{tabular}{llcccccc}
\toprule
\multirow{3}{*}{Dataset} & \multirow{3}{*}{Methods}
& \multicolumn{4}{c}{Video Depth} & \multicolumn{2}{c}{Camera} \\
\cmidrule(lr){3-6}\cmidrule(lr){7-8}
& & \multicolumn{2}{c}{scale} & \multicolumn{2}{c}{scale\&shift}
& \multirow{2}{*}{AUC@15$\uparrow$} & \multirow{2}{*}{AUC@30$\uparrow$} \\
\cmidrule(lr){3-4}\cmidrule(lr){5-6}
& & Abs Rel$\downarrow$ & $\delta<1.25\uparrow$ & Abs Rel$\downarrow$ & $\delta<1.25\uparrow$ & & \\
\midrule
\multirow{5}{*}{OmniVideo}
& freeze-C(ours)        & \textbf{0.181} & \textbf{0.792} & \textbf{0.113} & \textbf{0.876} & {89.132} & {93.945} \\
& freeze-D        & 0.193 & 0.761 & 0.118 & 0.869 & 89.281 & 93.983 \\
& freeze-A        & 0.263 & 0.672 & 0.157 & 0.800 & 82.531 & 89.174 \\
& freeze-C\&D     & 0.197 & 0.752 & 0.116 & 0.873 & 88.468 & 93.517 \\
& train-all       & 0.185 & 0.773 & 0.115 & 0.872 & \textbf{89.796} & \textbf{94.260} \\
\bottomrule
\end{tabular}
}
\end{table}

\begin{table}
\centering
\small
\setlength{\tabcolsep}{4pt}
\renewcommand{\arraystretch}{1.15}
\caption{\textbf{OmniVideo counterpart of the supervision form ablation.}
$+\mathcal{L}_{\text{feat}}$ adds a feature matching loss.}
\label{tab:supp_ablation_loss_omnivideo}
\resizebox{1.0\columnwidth}{!}{
\begin{tabular}{llcccccc}
\toprule
\multirow{3}{*}{Dataset} & \multirow{3}{*}{Mode}
& \multicolumn{4}{c}{Video Depth} & \multicolumn{2}{c}{Camera} \\
\cmidrule(lr){3-6}\cmidrule(lr){7-8}
& & \multicolumn{2}{c}{scale} & \multicolumn{2}{c}{scale\&shift}
& \multirow{2}{*}{AUC@15$\uparrow$} & \multirow{2}{*}{AUC@30$\uparrow$} \\
\cmidrule(lr){3-4}\cmidrule(lr){5-6}
& & Abs Rel$\downarrow$ & $\delta<1.25\uparrow$ & Abs Rel$\downarrow$ & $\delta<1.25\uparrow$ & & \\
\midrule
\multirow{2}{*}{OmniVideo}
& $+\mathcal{L}_{\text{feat}}$ & 0.182 & 0.783 & 0.114 & 0.873 & \textbf{89.146} & 93.840 \\
& base                         & \textbf{0.181} & \textbf{0.792} & \textbf{0.113} & \textbf{0.876} & 89.132 & \textbf{93.945} \\
\bottomrule
\end{tabular}
}
\end{table}
\section{Additional Limitation}
\label{sec:supp_limitation}
A limitation of our framework is that it relies on the base model having non-trivial predictive ability on the training videos. Because the method improves the model using self-generated pseudo targets, it cannot bootstrap effectively when the initial predictions are severely degraded. In such cases, the pseudo supervision may be dominated by errors, making the self-improving loop unstable or ineffective. Thus the proposed framework is most effective when initialized from a pretrained model with a reasonable geometric prior.
Consequently, the effectiveness of self-improving depends on the quality of the starting point, and the framework is better viewed as a post-training refinement mechanism than as a way to learn geometry from scratch.

\begin{figure*}
    \centering
    \includegraphics[width=\textwidth]{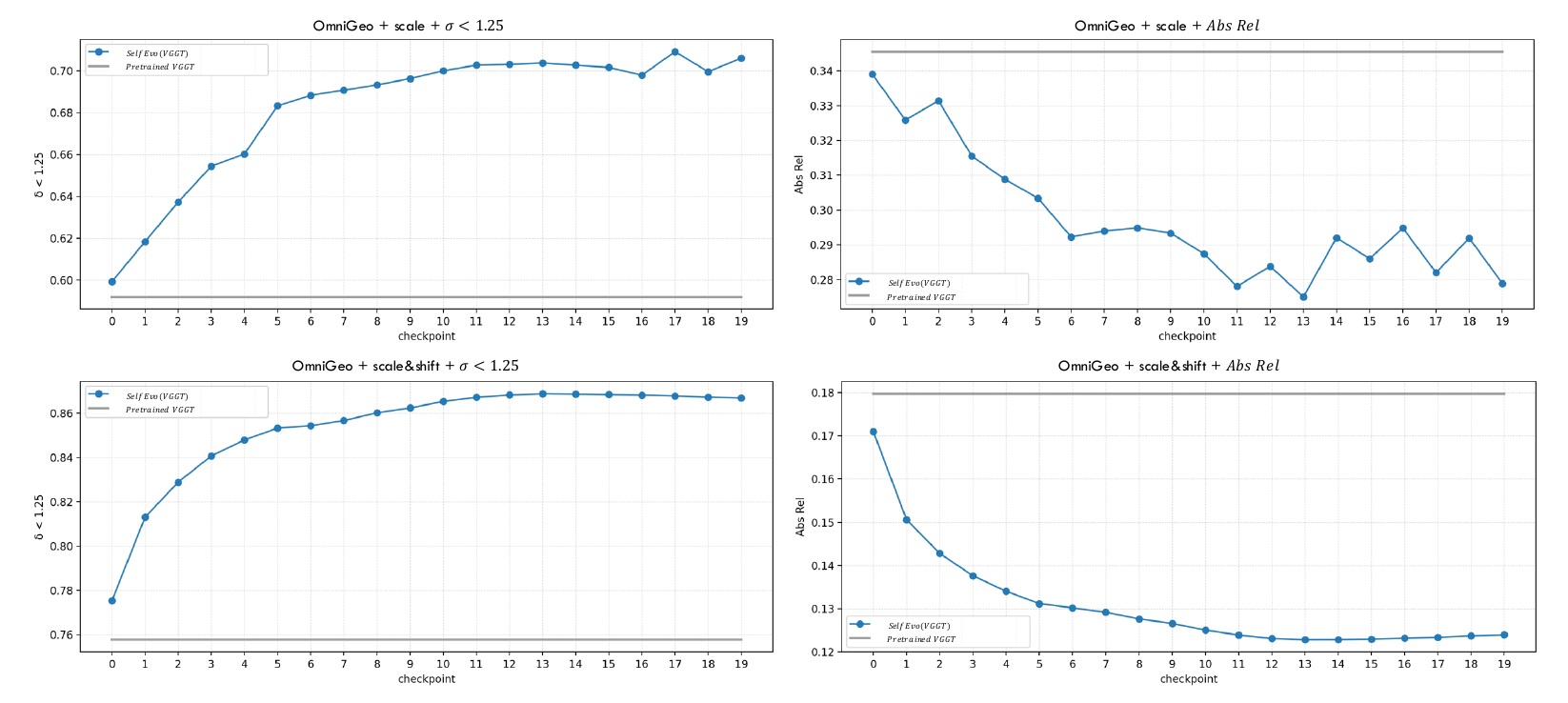}
    \caption{\small \textbf{Checkpoint-wise depth evaluation on the held-out OmniGeo benchmark.}
    We retrospectively evaluate all training checkpoints of \textsc{SelfEvo(VGGT)} on OmniGeo after training is completed. 
    Top row: \emph{scale} alignment; bottom row: \emph{scale-and-shift} alignment. 
    Left: $\delta < 1.25$ (higher is better); right: Abs Rel (lower is better). 
    The horizontal gray line denotes the pretrained \textsc{VGGT} baseline. 
    Across both alignment settings, self-improvement leads to substantial gains over the pretrained model throughout training, showing self-improvement on held-out depth benchmarks.}
    \label{fig:supp_depth_omnigeo}
\end{figure*}

\begin{figure*}
    \centering
    \includegraphics[width=\textwidth]{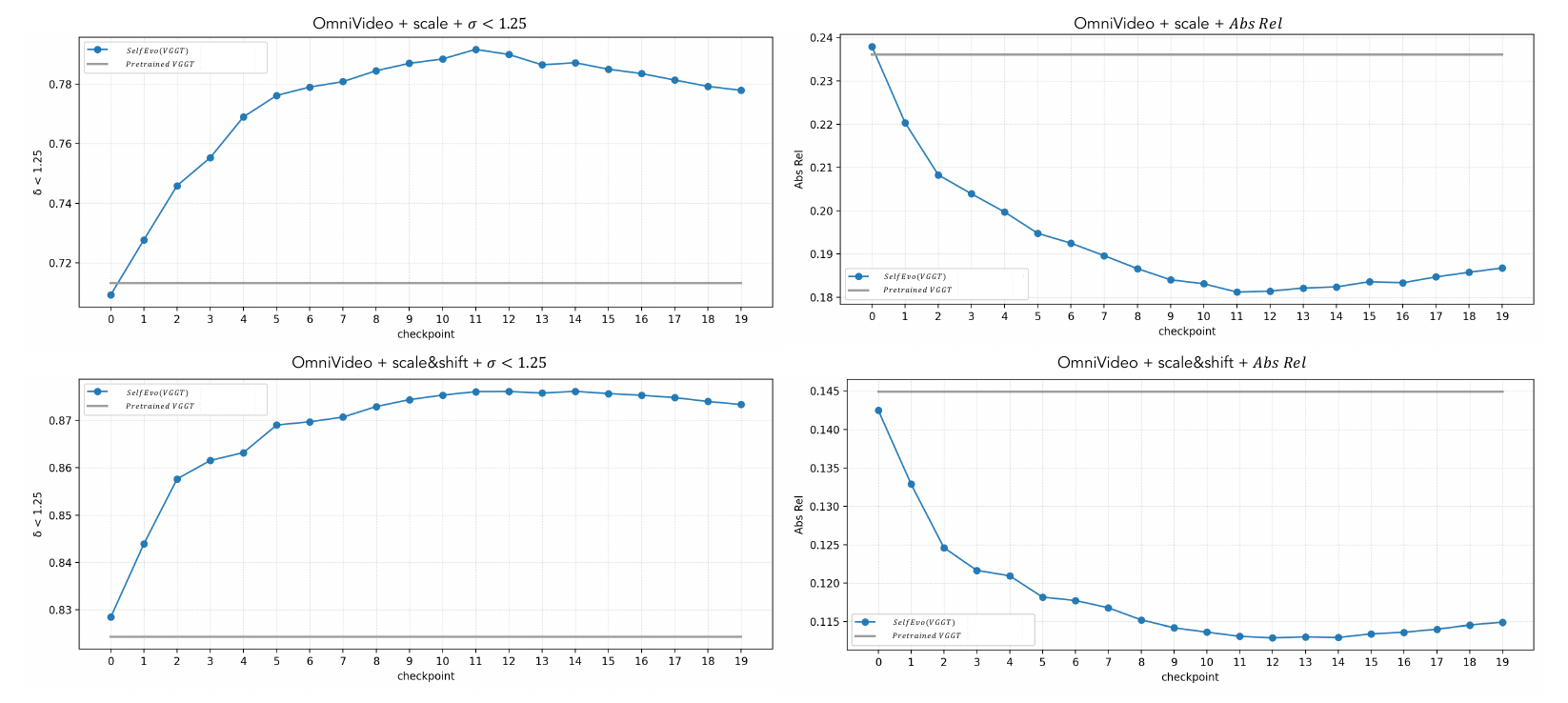}
    \caption{\small \textbf{Checkpoint-wise depth evaluation on the held-out OmniVideo benchmark.}
    We retrospectively evaluate all training checkpoints of \textsc{SelfEvo(VGGT)} on OmniVideo after training is completed. 
    Top row: \emph{scale} alignment; bottom row: \emph{scale-and-shift} alignment. 
    Left: $\delta < 1.25$ (higher is better); right: Abs Rel (lower is better). 
    The horizontal gray line denotes the pretrained \textsc{VGGT} baseline. 
    The curves show clear and sustained improvement across checkpoints, with rapid gains in the early stage and mild saturation later, again supporting self-improvement on depth benchmarks.}
    \label{fig:supp_depth_omnivideo}
\end{figure*}

\begin{figure*}
    \centering
    \includegraphics[width=\textwidth]{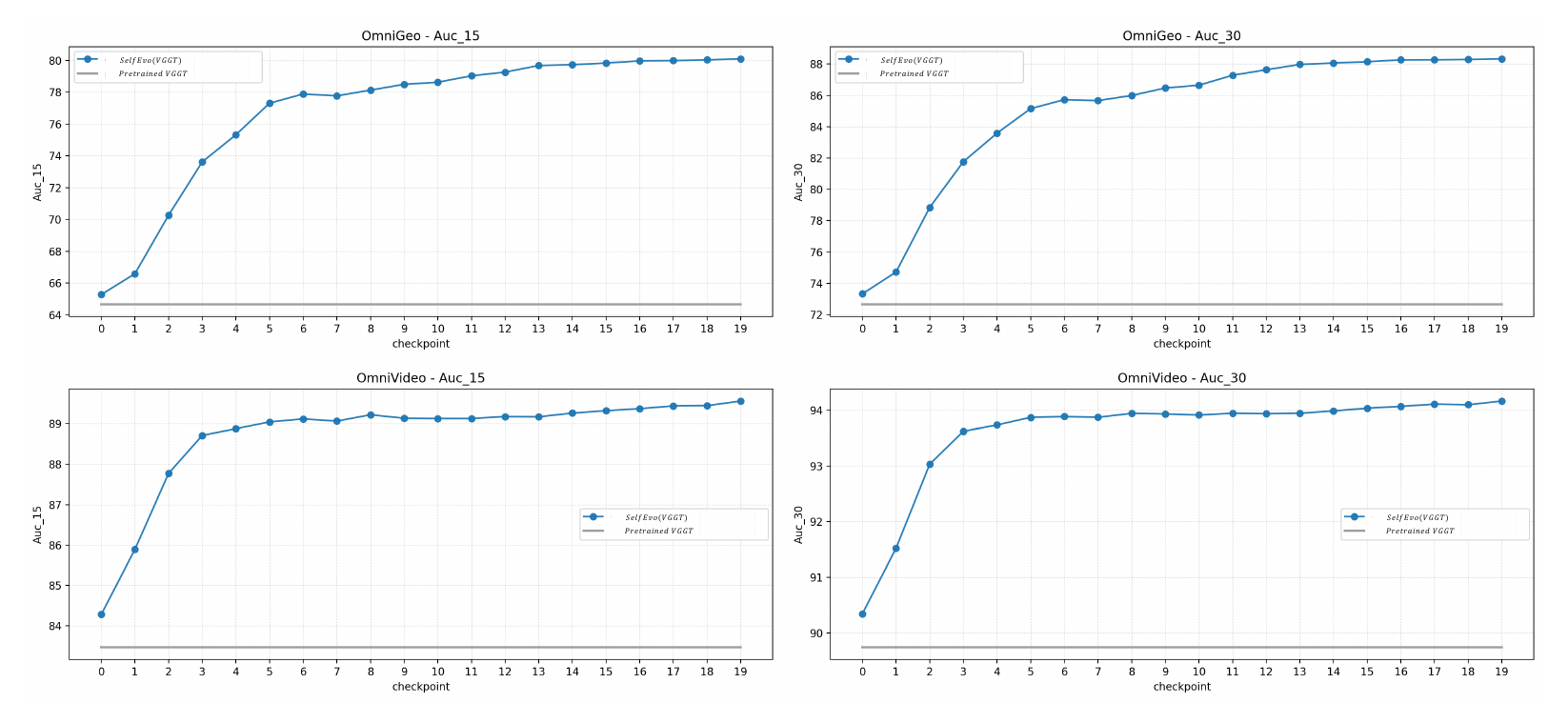}
    \caption{\small \textbf{Checkpoint-wise camera evaluation on held-out OmniGeo and OmniVideo benchmarks.}
    We retrospectively evaluate all training checkpoints of \textsc{SelfEvo(VGGT)} after training is completed. 
    Top row: OmniGeo; bottom row: OmniVideo. 
    Left: AUC@15; right: AUC@30. 
    Higher is better for all metrics. 
    The horizontal gray line denotes the pretrained \textsc{VGGT} baseline. 
    Camera estimation improves consistently throughout training on both held-out benchmarks, further demonstrating the self-improvement ability of our framework beyond depth prediction alone.}
    \label{fig:supp_camera_benchmarks}
\end{figure*}

\section{Self-Improvement on Held-Out Benchmarks}
\label{sec:supp_continuous_self_improvement}

We further examine whether our framework exhibits {self-improvement} throughout training. To this end, after training is completed, we retrospectively evaluate every checkpoint in our basic setting, using {VGGT} as the base model and OmniWorld-Game videos as unlabeled training data. We report checkpoint-wise performance on held-out OmniGeo and OmniVideo benchmarks. 

Fig.~\ref{fig:supp_depth_omnigeo} and Fig.~\ref{fig:supp_depth_omnivideo} show the depth results on the two held-out benchmarks. Across both datasets and both alignment protocols, the model improves steadily over the course of self-improvement: $\delta < 1.25$ increases substantially, while Abs Rel decreases substantially, relative to the pretrained baseline. The gains are especially pronounced in the early and middle stages of training, followed by saturation and mild fluctuations at later checkpoints. Nevertheless, the overall trend remains consistently positive across all depth metrics.

Fig.~\ref{fig:supp_camera_benchmarks} shows the camera results on the same held-out benchmarks. Both AUC@15 and AUC@30 improve throughout training on OmniGeo and OmniVideo, with rapid gains in the early stage and more gradual improvement afterward. The trends are largely monotonic and remain clearly above the pretrained baseline across all checkpoints.

Overall, these checkpoint-wise evaluations provide direct evidence that our framework enables \emph{continuous self-improvement}: as optimization proceeds, the model improves not only at the final checkpoint, but throughout training on held-out benchmarks, across both depth and camera estimation.

\begin{figure*}[t]
    \centering
    \includegraphics[width=\textwidth]{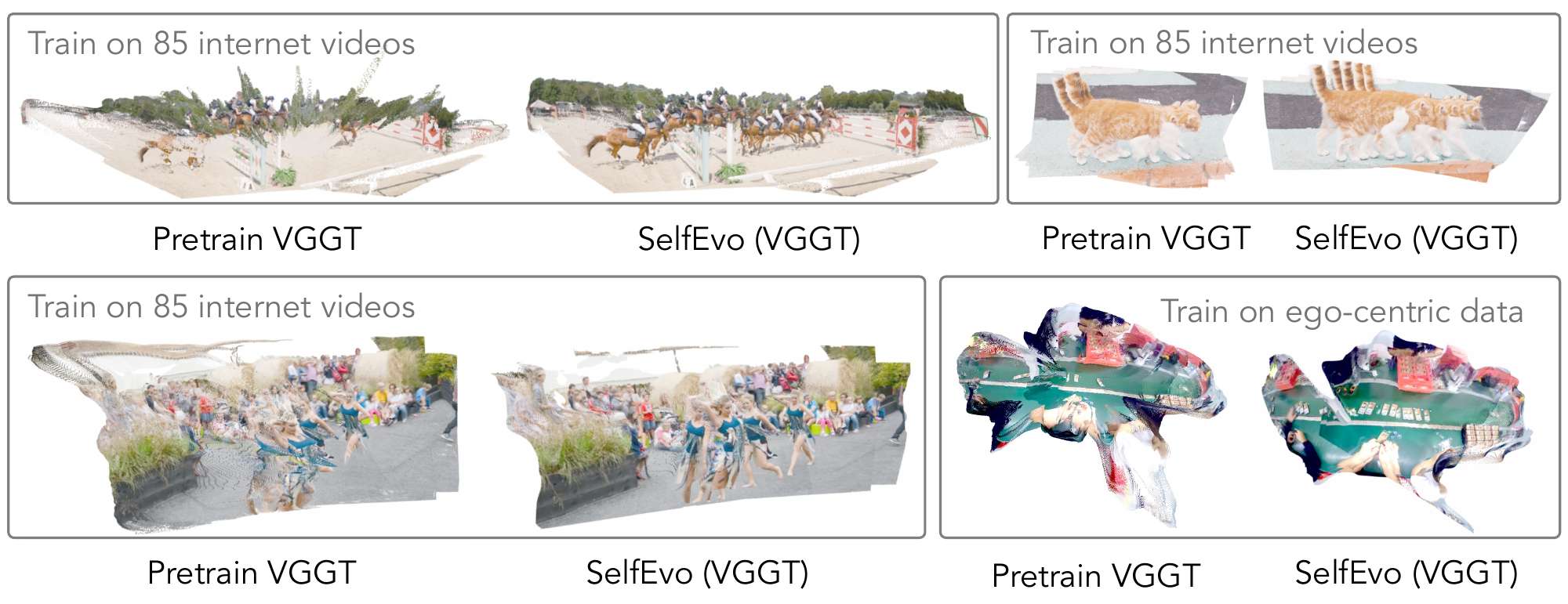}
    \caption{\small \textbf{Qualitative results on additional training sources.}
    We further apply \textsc{SelfEvo} to \textsc{VGGT} using two additional unlabeled video sources beyond our main setup: DynPose-100K and Egocentric-10K. 
    The first three examples show results after training on a very small fully in-the-wild subset of DynPose-100K, using only 85 internet videos, while the last example shows results after training on egocentric data. 
    In each pair, the left result is from the pretrained \textsc{VGGT} model and the right result is from \textsc{SelfEvo(VGGT)}. 
    }
    \label{fig:supp_additional_data_qualitative}
\end{figure*}

\section{Additional Training Data Sources}
\label{sec:supp_additional_data}

To further examine the generality of our self-improving framework, we also conduct additional training experiments on two video sources beyond the main setup: Egocentric-10K~\cite{buildaiegocentric10k2025} and DynPose-100K~\cite{rockwell2025dynpose}. In this section, we present qualitative results from these additional training runs.

\vspace{0.5em}
\noindent\textbf{Training on Egocentric-10K~\cite{buildaiegocentric10k2025}.}
We additionally train \textsc{VGGT} on Egocentric-10K, an in-the-wild egocentric video dataset characterized by strong hand visibility and dense active manipulation. Compared with earlier in-the-wild egocentric datasets, it contains richer hand-object interaction patterns and more frequent manipulation events, making it a challenging testbed for self-improving. Since egocentric videos are often captured with fish-eye cameras, we undistort all RGB videos before training. Despite the distinctive characteristics of this domain, including large ego-motion, frequent hand occlusions, and strong first-person viewpoint bias, our framework remains effective and yields promising qualitative improvements.

\vspace{0.5em}
\noindent\textbf{Training on DynPose-100K~\cite{rockwell2025dynpose}.}
We also train \textsc{VGGT} and \textsc{$\pi^3$} on DynPose-100K, a large-scale collection of diverse dynamic videos curated from internet sources, representing a fully in-the-wild training setting. Compared with our main setup, this data source is broader, less controlled, and more reflective of real-world video distributions. Notably, in our experiment we use only a very small subset of DynPose-100K, consisting of just 85 internet videos. Even with such limited in-the-wild training data, self-improvement still leads to clear qualitative improvements over the pretrained model, producing visually more coherent geometry and cleaner structure in dynamic scenes. This suggests that our framework remains effective even in small-scale, fully in-the-wild adaptation settings.

Overall, these additional experiments show that our framework is not restricted to a single unlabeled video source. It transfers across substantially different domains, including egocentric videos and fully in-the-wild internet videos. Notably, the DynPose-100K experiment suggests that effective self-improvement is possible even when training on only a very small in-the-wild subset. Qualitative examples from both settings are shown in Fig.~\ref{fig:supp_additional_data_qualitative}.

\end{document}